\documentclass{bmvc2k}
\usepackage{mathtools}

\title{Moving Object Segmentation in Jittery Videos by Stabilizing Trajectories Modeled in Kendall's Shape Space}

\addauthor{Geethu Miriam Jacob}{geethumiriam@gmail.com}{1}
\addauthor{Sukhendu Das}{http://www.cse.iitm.ac.in/~sdas/}{1}

\addinstitution{
Visualization and Perception Lab\\
Dept. of CS\&E\\
Indian Institute of Technology Madras\\
Chennai, India}

\runninghead{Jacob, Das}{Jittery Video Segmentation}

\def\etal{\emph{et al}\bmvaOneDot}

\begin{document}

\maketitle

\begin{abstract}
Moving Object Segmentation is a challenging task for jittery/wobbly videos. For jittery videos, the non-smooth camera motion makes discrimination between foreground objects and background layers hard to solve. While most recent works for moving video object segmentation fail in this scenario, our method generates an accurate segmentation of a single moving object. The proposed method performs a sparse segmentation, where frame-wise labels are assigned only to trajectory coordinates, followed by the pixel-wise labeling of frames. The sparse segmentation involving stabilization and clustering of trajectories in a 3-stage iterative process. At the 1st stage, the trajectories are clustered using pairwise Procrustes distance as a cue for creating an affinity matrix. The 2nd stage performs a block-wise Procrustes analysis of the trajectories and estimates Frechet means (in Kendall's shape space) of the clusters. The Frechet means represent the average trajectories of the motion clusters. An optimization function has been formulated to stabilize the Frechet means, yielding stabilized trajectories at the 3rd stage. The accuracy of the motion clusters are iteratively refined, producing distinct groups of stabilized trajectories. Next, the labels obtained from the sparse segmentation are propagated for pixel-wise labeling of the frames, using a GraphCut based energy formulation. Use of Procrustes analysis and energy minimization in Kendall's shape space for moving object segmentation in jittery videos, is the novelty of this work. Second contribution comes from experiments performed on a dataset formed of 20 real-world natural jittery videos, with manually annotated ground truth. Experiments are done with controlled levels of artificial jitter on videos of SegTrack2 dataset. Qualitative and quantitative results indicate the superiority of the proposed method.
\end{abstract}

\section{Introduction}
\label{sec:intro}
Moving Object Segmentation aims to segment a moving object, by classifying each pixel as belonging to either foreground or background. Moving Object Segmentation is a pre-processing step for high level applications such as  video retrieval, video editing and activity recognition. Existing approaches segment the moving objects from either static videos~\cite{mittal2009scene,barnich2011vibe,mittal2004motion,stauffer1999adaptive,karsch2012depth} or moving camera videos~\cite{ma2012maximum,lee2011key,ochs2014segmentation,zhang2013video,Papazoglou_2013_ICCV}. However, very few works explore the performance of the segmentation algorithm in wobbly/jittery videos. Jittery videos are taken from handheld devices, such as camcorders or cell phones, available ubiquitously with the users. Our method proposes to accurately segment a single moving object from jittery videos. 

Moving object segmentation can be performed using unsupervised as well as semi-supervised methods. Trajectory based~\cite{shi2000normalized, elhamifar2009sparse,jain2014supervoxel, brox2010object,ochs2011object,ochs2014segmentation} and graph based~\cite{lee2011key,ma2012maximum,Papazoglou_2013_ICCV,zhang2013video,faktor2014video,wang2015saliency} methods are popular in unsupervised techniques. Both the methods depends heavily on the accuracy of optical flow estimation and the distinctiveness of the motion between the moving objects and background. Hence, when the video is jittery, they fail due to the reduced distinctiveness in the motion between moving objects and background. Semi-supervised techniques~\cite{li2013video, avinash2014seamseg,perazzi2015fully,marki2016bilateral,tsai2016video} take user input in certain frames and the labels are propagated to the rest of the frames. This helps in tracking the moving object more precisely than the unsupervised techniques, even if the video is jittery in nature. But, when the video is long, much user input is required, which is undesirable.

We assume that there is a single moving object present in all the frames. Our method is an unsupervised trajectory based approach, which adopts the idea of stabilization to generate accurate clusters of trajectories. Given a jittery video as input, the method extracts the trajectories~\cite{sundaram2010dense}, stabilizes and clusters them separately as moving object and background trajectories. The stabilization aids in clustering the trajectories accurately. We propose a joint solution for the tasks of stabilization and segmentation, which needs less computational time than performing them separately. Also, the performance of the joint solution is better than executing the two tasks separately, since the quality of videos degrade after stabilization due to warping. We consider trajectories, as shapes and model them in Kendall's shape space~\cite{dryden1998statistical} for stabilization and clustering, thus assigning labels to the trajectory coordinates in video frames (sparse segmentation). The sparse segmentation involving stabilization and clustering of trajectories in a 3-stage iterative process. First, the trajectories are clustered using pairwise Procrustes distance as a cue for creating an affinity matrix. Second stage performs a block-wise Procrustes analysis of the trajectories and estimates Frechet means (in Kendall's shape space) of the clusters. The Frechet means represent the average trajectories of the motion clusters. An optimization function has been formulated to stabilize the Frechet means, yielding stabilized trajectories at the third stage. The stabilized trajectories produce enriched clusters of motion paths for the next iteration. The accuracy of the motion clusters are iteratively refined, producing distinct groups of stabilized trajectories. Next, the labels obtained from the sparse segmentation are propagated for pixel-wise labeling of the frames, using a GraphCut based energy formulation.

The contributions of the work are as follows: 1) Procrustes distance as a similarity measure between trajectories for computing the affinity matrix for clustering, 2) Use of Procrustes analysis and a cost (energy) function for minimization in Kendall's shape space for the purpose of stabilization of trajectories, and 3) Developing a dataset of 20 real-world natural jittery videos with manual annotation of ground truth by us, for performance analysis. The superiority of the proposed method is shown on two datasets: a) 20 natural jittery videos and b) artificial jitter fused on videos of a standard (stable camera motion) segmentation dataset, SegTrack2~\cite{segdataset}. It is shown experimentally that the method works well for both stable as well as jittery videos. Our method generates accurate clusters in shape space, where many conventional trajectory clustering algorithms fail due to the randomness of the jittery motion. The better result of the proposed method is due to the stabilization of the trajectories, which makes the non-smooth foreground and background trajectories distinct.

The paper is organized as follows. Section 2 gives a brief review of work done on video stabilization and moving object segmentation. Section 3 describes the approach of trajectory stabilization and clustering in Kendall's shape space, whereas section 4 discusses about the procedure followed for pixel-wise labeling of frames from the sparse segments. Section 5 gives details of experiments and results obtained, while Section 6 concludes the paper.

\section{Related Work}
 A brief review of various stabilization and segmentation methods are presented below:

\textbf{Video Stabilization:}
Video stabilization is the process of smoothing the videos which contain considerable amount of wobble/jitter. Gaussian smoothing of neighboring transformations~\cite{matsushita2006inpaint} and low pass filtering of trajectories in subspace~\cite{liu2011subspace,liu2013joint} are some of the methods developed for video stabilization. The works in~\cite{chen2008capturing,Gleicher2008cinema}
generates the smooth path by fitting curves on the camera
path. Meanwhile, several works
pose the problem of video stabilization as an optimization
function. The method proposed by Grundmann~\etal~\cite{GrundmannKwatra2011} obtains the desired camera path by posing a Linear
Programming Problem (LPP). Liu~\etal proposed variational methods known as bundled paths model~\cite{liu2013bundled} based on camera paths and steadyflow model~\cite{liu2014steadyflow} based on optical flow. All the above methods perform smoothing or optimizations on all the trajectories or optical flow paths separately, due to which the stabilization may not be consistent across trajectories. The method proposed recently in~\cite{Jacob2016VSP} estimates a mean trajectory to represent and stabilize the camera motion. Our method is inspired from this work and aims at stabilizing and estimating the mean trajectories for both foreground and background motions. 

\textbf{Video Object Segmentation:}
 The first kind of trajectory based segmentation method was introduced in~\cite{shi1998motion}, where motion profiles are segmented using normalized cuts~\cite{shi2000normalized}. Subspace clustering methods~\cite{elhamifar2009sparse,rao2008motion,liu2010robust,soltanolkotabi2014robust} aims at clustering trajectories by modeling the trajectory matrix in subspaces. Brox~\etal~\cite{brox2010object} proposed the concept of long term trajectories from optical flow. The long term trajectories track the pixels across frames. The clusters of the trajectories are converted to object segments using a variational model in~\cite{ochs2011object}. The authors proposed an alternate clustering of long term trajectories using hypergraphs in~\cite{ochs2012higher}. Fragkiadaki~\etal~\cite{fragkiadaki2012video} cluster trajectories by computing discontinuities between trajectories that are Delaunay neighbors. The method in~\cite{chen2015video} uses global and local trajectory affinities to create clusters and later performs a GraphCut based energy minimization for generating segments. Potts' model was used in~\cite{ochs2014segmentation} to generate segments of objects. Recently, Liu~\etal~\cite{chen2015video} proposed a method to use both local and global affinities for spectral clustering, whereas the rest of the methods use local affinities.

Apart from trajectory based methods, many recent techniques propose novel unary and pairwise potentials for the GraphCut minimization. Lee~\etal~\cite{lee2011key} ranks proposals in terms of appearance, objectness and motion and then perform spatio-temporal segmentation on the top ranked proposals. In~\cite{zhang2013video}, a
layered directed acyclic graph based optimization is performed on the object
proposals to obtain the segmented objects, whereas in~\cite{wang2015saliency}, a saliency based proposal for unary potential is introduced. A two stage approach is proposed
in~\cite{Papazoglou_2013_ICCV}, where the first stage produces an initial estimate of the object using
inside-outside maps and the second stage refines the segmentation. A non local consensus voting on superpixels was performed in~\cite{faktor2014video} to obtain the segmented objects. None of the above methods have been shown to work well for jittery videos. Our method is an unsupervised approach, which utilizes affinities of trajectories in Kendall's shape space, defined using the Procrustes distance. 

\begin{figure}
\begin{center}
\includegraphics[scale=0.39]{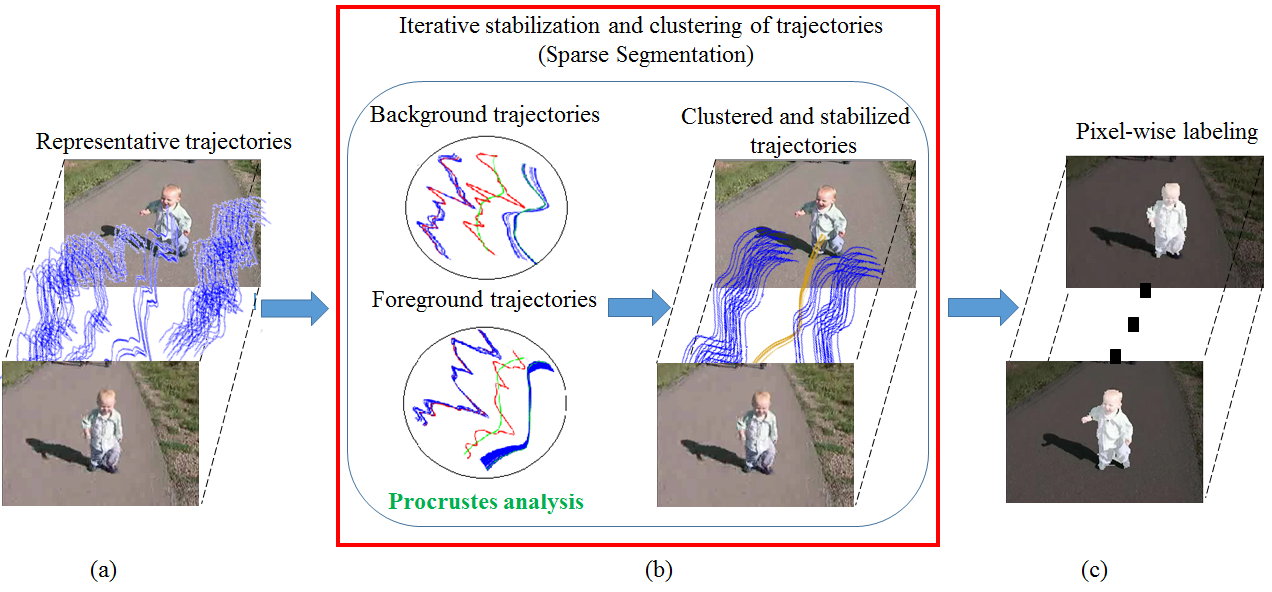}
\end{center}
\caption{Proposed Framework: (a) Generate the representative trajectories and model them in Kendall's shape space, (b) iterative stabilization and clustering using Procrustes Analysis of both foreground and background trajectories and (c) Pixel-wise labeling of frames.}
\label{framework}
\end{figure}
\section{Proposed Framework}
The proposed framework is shown pictorially in Figure~\ref{framework}. The stages of our framework are:
\begin{enumerate}
    \item \textit{Model 'representative trajectories' in pre-shape space}: Point trajectories~\cite{sundaram2010dense} are extracted and divided into blocks. For each block, representative trajectories are computed and modeled in Kendall's pre-shape space.
    \item \textit{Iterative trajectory stabilization and clustering (sparse segmentation):} The trajectories are stabilized and clustered iteratively using Procrustes analysis to obtain motion clusters. The stabilization of trajectories aids in producing well formed clusters. 
    \item \textit{Pixel-wise labeling of frames:} The labels (of clusters) assigned to the trajectory coordinates are taken as input and a GraphCut based MRF formulation is applied to obtain the pixel-wise labeling. Appearance and location prior from superpixels of the clustered trajectory coordinates are taken as cues for the energy minimization. 
\end{enumerate}
  
\subsection{Representative Trajectories as pre-shapes}
Point trajectories are extracted using~\cite{sundaram2010dense} and the frames are divided into varying-sized blocks based on the number of trajectories spanning the whole block. Empirically, the block size is selected such that the minimum number of trajectories spanning the entire block should be at least 10\% of the total number of trajectories extracted. This ensures that the trajectories belonging to both foreground and background are included in the further processing. From the set of trajectories spanning an entire block, one trajectory is selected for each superpixel~\cite{achanta2012slic} in the first frame. Let N be the number of frames in the block and K be the number of representative trajectories selected. The $k^{th}$ trajectory is represented in the matrix form as:

\begin{equation}
  X^k = \begin{bmatrix}
x_1^k & x_2^k & \ldots & x_N^k \\
y_1^k & y_2^k & \ldots & y_N^k
\end{bmatrix}^T 
\end{equation}

Here, $X^k$ is the trajectory matrix, $x_i^k$ and $y_i^k$ represents the x and y coordinates of the $k^{th}$ trajectory in $i^{th}$ frame of the block.

As discussed in~\cite{dryden1998statistical}, a shape is defined as the geometrical information of a sequence
of an object that is invariant under translation, scale
and rotation transformations. A shape space is the set of all possible such shapes. Transforming the trajectory coordinates to the shape space ensures that they are aligned geometrically before any processing. First, the pre-shape configuration of the trajectory is computed. Pre-shape space, the space of all pre-shapes, is a precursor to the shape space, where the configurations are invariant to location and scale. The pre-shape of the trajectory $X^k$ is given as:
\begin{equation}\label{preshape}
Z^k=\frac{C.X^k}{||C.X^k||_F},  
\end{equation}
where '.' is the matrix multiplication function, $Z^k$ is the pre-shape of the $k^{th}$ configuration and $C=I_N-\frac{1}{N}1_N1_N^T$. $I_N$ is the $N\times N$ identity matrix and $1_N$ is the $N\times 1$ vector of ones. The original trajectory configurations are centered and scaled to be of unit norm. Thus, the preshape space is a hypersphere of 2(N-1) dimensions (see~\cite{dryden1998statistical} for more details). 
\begin{figure}
\includegraphics[height=40mm,width=127mm]{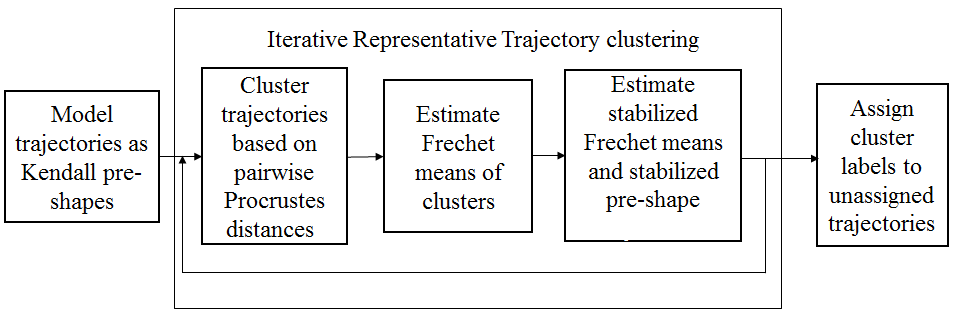}
\caption{ Iterative estimation of trajectory clusters: The process of clustering the trajectories, estimating the Frechet means of the clusters and stabilizing them to obtain the stabilized trajectories are performed in an iterative manner. Each iteration refines the clusters. Later, unassigned trajectories are assigned cluster label based on their similarity with the estimated Frechet means.}
\label{flowchart}
\end{figure}

\subsection{Iterative Stabilization and Clustering of Trajectories (Sparse Segmentation)}
The flowchart of the iterative procedure for stabilization and clustering of trajectories to obtain sparse motion segments is shown in Figure~\ref{flowchart}. Pairwise Procrustes distances between the representative trajectories are used to generate the affinity matrix for clustering, whereas Generalized Procrustes analysis (GPA) of the representative trajectories is performed to estimate and stabilize the motion of the foreground and background. Following are the three main stages for obtaining the clusters of trajectories: 

\subsubsection{Cluster trajectories using pairwise Procrustes distances}
Pairwise Procrustes distances between the representative trajectories are considered for creating the affinity matrix for clustering. Trajectories belonging to the same cluster move together and hence should have higher affinity. Procrustes distance between two trajectories is defined as the distance between them in the shape space. The pre-shape configurations obtained from the previous stage are optimally rotated such that they become geometrically aligned. The transformation of the trajectories to the shape space makes them comparable. Let $Z^i$ and $Z^j$ be two trajectory configurations in the pre-shape space. The Procrustes distance between the pre-shapes is given as:
 \begin{equation}
d_{proc}(i,j)=||Z^i-Z^j\hat{\Gamma}||_2 
\end{equation}
where $\hat{\Gamma}=argmin_{\Gamma\in SO(2)}||Z^i-Z^j\Gamma||_2$. Here $\hat{\Gamma}=UV^T$, where ${Z^{i}}^{T} Z^j=V\Lambda U^T$ is the optimal rotation that aligns $Z^i$ and $Z^j$, using the Singular Value Decomposition (SVD). This distance compares the shapes of the trajectories, which discursively tells the difference in the motion of the trajectories across the frames of the block. Lower the distance, the more similar their motion is.

The affinity matrix, which defines the affinity between the trajectories, for performing spectral clustering is created based on the pairwise Procrustes distance. The elements of the affinity matrix, $Afy$ of size $K \times K$ is given as:
\begin{equation}
Afy(i,j)=exp(-\frac{d_{proc}(i,j)}{\omega})
\end{equation}

The parameter $\omega$ is empirically estimated as 0.02 for optimal performance. Spectral clustering~\cite{von2007tutorial} is performed on the affinity matrix ($Afy$) created. This generates foreground and background clusters of trajectories. Thus, each trajectory is assigned a cluster label.

\subsubsection{Estimate Frechet means of clusters}\label{cluster}
Generalized Procustes analysis (GPA) as discussed in~\cite{dryden1998statistical} is performed on each of the trajectory cluster. In GPA, the clusters are aligned in the shape space such that the estimates of the motion of the clusters is obtained. The pre-shape configurations are optimally rotated such that the sum of distances between the trajectories of the cluster in the shape space is minimized. Let the indicator function $I_i^m$ for $i^{th}$ pre-shape trajectory denote whether the trajectory belong to the $m^{th}$ cluster or not and let $\mathcal{K}^m$ be the number of trajectories in the $m^{th}$ cluster, i.e $\mathcal{K}^m=\sum_{i=1}^KI_i^m$. Optimal rotations required for aligning the configurations are estimated using the following optimization function:

\begin{equation} \label{Procrustes}
\begin{split}
\{\hat{\Gamma}_i^m\} & = \underset{\Gamma_i^m\in SO(2)}{arg min}\frac{1}{\mathcal{K}^m}  \sum_{i=1}^K\sum_{j=i+1}^KI_i^mI_j^m\bigg[||Z^i\Gamma_i^m-Z^j\Gamma_j^m||_2\bigg]^2  \forall m \\
&=
\underset{\Gamma_i^m\in SO(2)}{arg min}\frac{1}{\mathcal{K}^m}  \sum_{i=1}^KI_i^m\bigg[||Z^i\Gamma_i^m-\frac{1}{\mathcal{K}^m}\sum_{j=1}^KI_j^mZ^j\Gamma_j^m||_2\bigg]^2
\end{split}
\end{equation}
where, the rotation matrices $\{\Gamma_i^m\}$ are constrained to belong to the Special Orthogonal group of dimension 2 (SO(2)), $\{\hat{\Gamma}_i^m\}$ are the optimal rotation matrices aligning the trajectories of each cluster together and the factor [$\frac{1}{\mathcal{K}^m}\sum_{j=1}^KI_j^mZ^j\Gamma_j^m$] is the Procrustes estimate of the mean of the trajectories of $m^{th}$ cluster. Since, this work focuses on the foreground background separation, the number of clusters ($m$) is set as 2.

The optimization function in Eqn~\ref{Procrustes} returns a set of transformations (rotation matrices, $\{\hat{\Gamma}_i^m\}, m=1,2$) for each cluster. The mean of the trajectories of the cluster in the shape space, popularly known as Frechet means, gives estimates of the motion of the clusters. 
\subsubsection{Estimate stabilized Frechet means and trajectories}
Frechet means give the mean trajectories of the clusters in the shape space. Let the Frechet mean of the $m^{th}$ cluster be given as $F_m=\frac{1}{\mathcal{K}^m}\sum_{j=1}^KI_j^mZ^j\hat{\Gamma}_j^m$. The Frechet means of the moving object and the background estimated are separately stabilized. The stabilization is performed using the optimization function:
\begin{equation}\label{stab}
\hat{F}_m^s=argmin_{F_m^s}\bigg[||F_m^s-F_m||_2\bigg]^2+\lambda\sum_{i=1}^N\bigg[||F_m^s(i,:)-\tilde{F}_m^s||_2\bigg]^2
\end{equation}
where $\lambda$ is the tuning parameter (empirically set as 0.2 for low jitter and 0.6 for high jitter), $F_m^s$ is the stabilized Frechet mean of the $m^{th}$ cluster, $\hat{F}_m^s$ is the optimal stabilized Frechet mean and $\tilde{F}_m^s=\frac{1}{N}\sum_{i=1}^NF_m^s(j,:)$. The first term in the function defined in Eqn~\ref{stab} keeps the stabilized Frechet mean close to the original and the second term, a variance minimization term, reduces the jitteriness in the mean trajectory. The function is solved using a jacobi iterative solver, by differentiating the functions to obtain a closed form solution. The solution to the optimization problem in Eqn~\ref{stab}, is:
\begin{equation}
(\hat{F}_m^s(r,:))^{t+1}=\alpha F_m(r,:)+\beta\sum_{i=1,i\neq r}^N(\hat{F}_m^s(i,:))^t
\end{equation}
where $\alpha=(1+\lambda-\frac{\lambda}{N})^{-1}$ and $\beta=\frac{\lambda}{N}\alpha$. Here, $t$ indicates the iteration index and is set to 5. Sparse segmentation is by itself an iterative process of alternating stabilization and clustering and hence number of iterations for the jacobi solver is set low. (Kindly refer the details given in App. A.1 in the supplementary file~\cite{KendallSegSupp} for derivation of the closed form solution.)

To obtain the stabilized trajectories, the stabilized Frechet means are transformed back to the pre-shape space. For $k^{th}$ trajectory, if $I_k^m=1$, the stabilized pre-shape trajectory is obtained from the stabilized Frechet mean $\hat{F}_m^s$, as $Z^k=\hat{F}_m^s\Gamma_k^{m^{T}}$. The three stages of sparse segmentation, namely clustering the shapes, Frechet mean estimation of the clusters and estimation of the stabilized pre-shape trajectories is an iterative procedure. Iteratively, the clusters become refined and accurate, and the Frechet means become more distinct. The number of iterations for obtaining the clusters is set as 3.


Once the iterative procedure ends, the trajectories which are not labeled and that span at least 70\% of the frames in the block are assigned a cluster label. The trajectory is assigned the label of the mean which has the lowest Procrustes distance with itself. Our method is more robust to non-smooth trajectories, due to stabilization process. Also, the method is less sensitive to outliers. If outliers are present, the pixel-wise foreground labeling process, as described in the following section, refines and generates accurate segments of foreground and background.

\begin{figure}
\begin{center}
\includegraphics[scale=0.32]{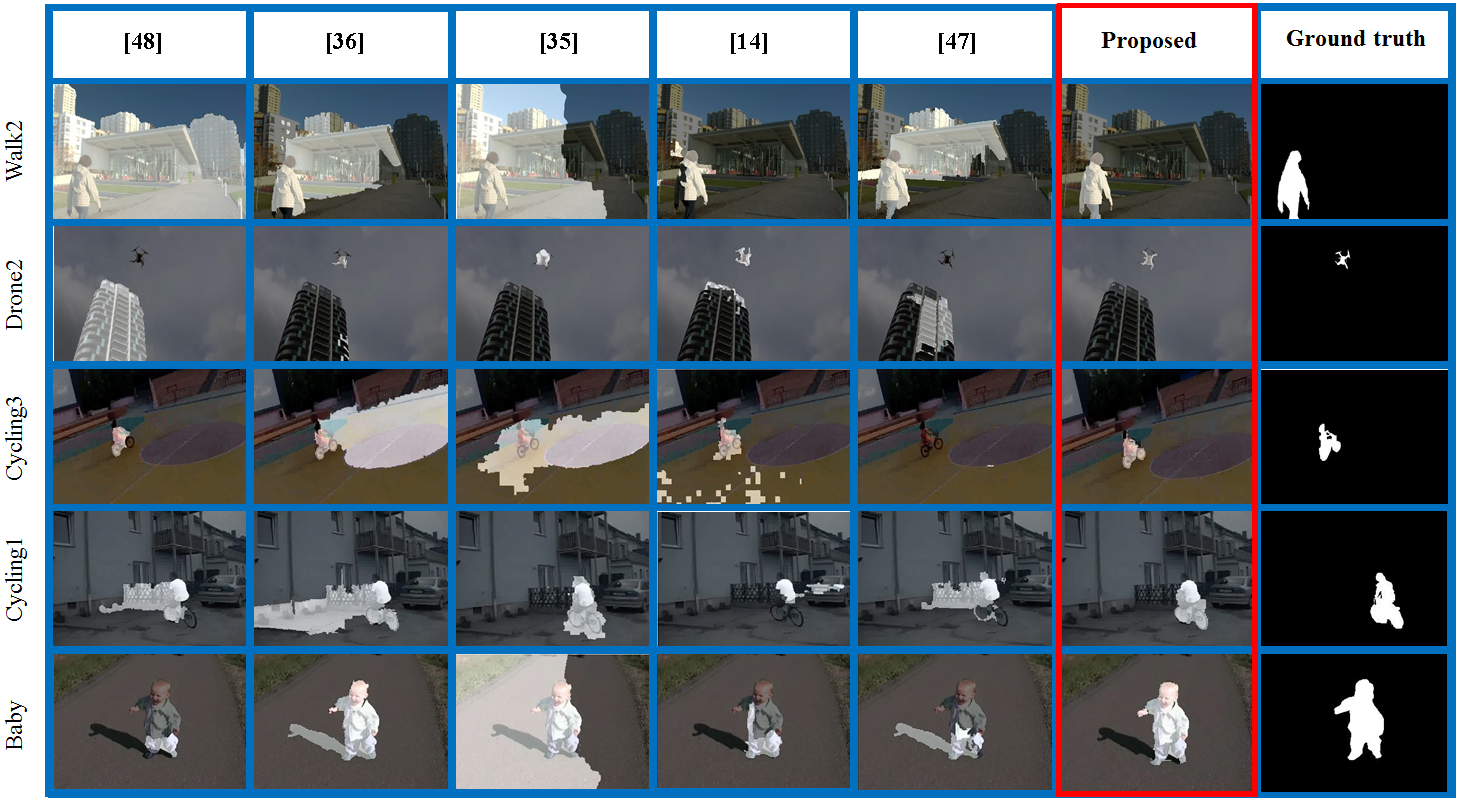}
\end{center}
\caption{Visual comparison of the performance of our proposed method of tracking, with five state-of-the-art methods~\cite{zhang2013video,Papazoglou_2013_ICCV,ochs2014segmentation,faktor2014video,wang2015saliency}, shown with one frame each of the video shots: \textit{Walk2, Drone2, Cycling3, Cycling1, Baby}. The column highlighted with the red template shows our result and the last column shows the ground truth of the frame.}
\label{comparison}
\end{figure}

\subsection{Pixel-wise labeling of frames}
Further, the superpixels~\cite{achanta2012slic} of the labeled trajectory coordinates in each frame are given the same label as that of the corresponding trajectories. MRF based optimization (as in~\cite{zhang2013video}) is performed on the superpixels of the frame, Here, an appearance and location based unary potential, along with a color induced pairwise potential is defined. The optimization is performed using GraphCut algorithm~\cite{rother2004grabcut}. This process classifies the superpixels in all frames as either foreground or background.

\section{Experimental Results}
Experiments were performed on two types of datasets: 20 real-world natural jittery videos; and a few videos from the standard segmentation dataset, SegTrack2, with artificial jitter fused in them to appear jittery. The proposed method is compared with 5 other state-of-the-art techniques~\cite{zhang2013video,Papazoglou_2013_ICCV,ochs2014segmentation,faktor2014video,wang2015saliency}, whose codes are publicly available. Only unsupervised methods are considered for the comparison with the proposed method. The measure used for performance evaluation is Intersection Over Union (IoU)~\cite{faktor2014video,li2013video,perazzi2015fully}, defined as the ratio of intersection to the union of the ground-truth and the segmented output.

\begin{table}
\begin{center}
\begin{tabular}{ |p{2.3cm}|p{1.1cm}|p{1.1cm}|p{1.1cm}|p{1.1cm}|p{1.1cm}|p{1.2cm}| }
\cline{2-7}
\multicolumn{1}{c|}{}&  \multicolumn{6}{c|}{\textbf{IoU scores of competing methods}}\\ \hline
\textbf{Video samples}& ~\cite{zhang2013video} & ~\cite{Papazoglou_2013_ICCV} & ~\cite{ochs2014segmentation}& ~\cite{faktor2014video}&~\cite{wang2015saliency}&Proposed\\
\hline
Walk2& 0.009&0.123&0&0.480&0.151&\textbf{0.841} \\ 
\hline
Cheery\_Girl& 0.144&0.201&0.09&0.587&0.573&\textbf{0.756} \\ \hline
Doll& 0.139&0.926&0.819&0.350&0.078&\textbf{0.933} \\ 
\hline
Baby& 0.116 &0.671&0.007&0.360&0.222&\textbf{0.847}\\ 
\hline
Skating1& 0.033&0.248&0.318&0.627&0.523&\textbf{0.713} \\
\hline
Climb1 & 0.54&0.764&0.03&\textbf{0.844}&0.476&0.810 \\
\hline
Drone2& 0.487&0.436&0.549&0.325&0.348&\textbf{0.588}\\ 
\hline
Train &  0.211&0.37&0.535&0.837&0.831&\textbf{0.850} \\ 
\hline
Cycling1& 0.558&0.359&0.610&0.462&0.342&\textbf{0.613} \\
\hline
Staircase2&0.875&0.889&0.801&0.001&0.103&\textbf{0.901} \\ 
\hline
\textbf{Average*}&0.425&0.431&0.392&0.529&0.421&\textbf{0.723}\\
\hline
\textbf{\textbf{Single object?}}& Y &Y & N &N &Y &Y\\
\hline
\end{tabular}
\caption{Performance analysis of segmentation on a subset of the 20 real-world jittery videos in terms of IoU score (higher, the better). See App. A.2 of the supplementary material for the individual results of all 20 videos~\cite{KendallSegSupp}. Last row indicates whether the method handles single or multiple moving objects. * - Average over all real-world 20 videos. }
\label{tab1}
\end{center}
\end{table}
A dataset of 20 real-world natural jittery videos with moving object was formed, among which 7 videos are from stabilization datasets~\cite{stabdataset} (\textit{Walk1, Walk2, Cheery\_Girl, Doll, Baby, Skating1, Car}), 5 jittery videos from HMDB51 dataset~\cite{hmdbdataset} (\textit{Cycling1, Cycling2, Climb1, Climb2, Skating2}), 4 videos from the public resource, YouTube (\textit{Drone1, Drone2, Drone3, Train}) and 4 videos shot by us (\textit{Staircase1, Staircase2, Cycling3, Dog}) using a hand-held camcorder in motion. The ground truth of every fifth frame of each video was manually annotated by us. The dataset and the ground truth can be downloaded from our website~\cite{KendallSeg}. Figure~\ref{comparison} shows some qualitative results of all the five methods and ours on one frame on each of the videos: \textit{Walk2, Drone2, Cycling3, Cycling1} and \textit{Baby}. In all cases, unsatisfactory performance of prior work is visible. Our method performs the best for all the cases in the figure. Table~\ref{tab1} shows the quantitative results (on a subset of the 20 videos) for the experiments performed on natural jittery videos. As seen in the table, our method performs the best for all the videos except for one case, where ours is a very close second. On an average, as given in the second last row of table~\ref{tab1}, our method performs the best by quite a margin. The last row indicates if the comparing method segments single moving object or multiple moving objects. The iterative method of stabilization using Procrustes analysis contributes to this enhanced performance. The method takes 5-6 seconds per frame (of size $360 \times 640$), when averaged over 10 videos, on an i7, 2GHz 16GB machine.

\begin{table}[!h]
\begin{center}
 \begin{tabular}{ |p{2.3cm}|p{1.2cm}|p{1.2cm}|p{1.2cm}|p{1.2cm}| p{1.2cm}|p{1.2cm}|}
\cline{2-7}
\multicolumn{1}{c|}{}&  \multicolumn{6}{c|}{\textbf{IoU scores of competing methods}}\\ 
\hline
\textbf{Jitter Level ($\sigma$)}&~\cite{zhang2013video} & ~\cite{Papazoglou_2013_ICCV} & ~\cite{ochs2014segmentation}& ~\cite{faktor2014video}&~\cite{wang2015saliency}&Proposed\\
\hline
Low (0.05) & 0.586&0.637&0.327&0.692&0.535&\textbf{0.695} \\ 
\hline
Medium (0.15) & 0.551&0.5749&0.525&0.686&0.506&\textbf{0.690} \\ \hline
High (0.25) & 0.543&0.585&0.479&0.654&0.470&\textbf{0.688} \\ \hline
\end{tabular}
\caption{Comparison of the average (over 72 videos) segmentation performances for low, medium and high levels of jitter, fused into stable videos (8) of SegTrack2 dataset~\cite{segdataset}.}
\label{tab3}
\end{center}
\end{table}

Experiments were also done on videos with different artificial random jitter levels fused on video frames of SegTrack2~\cite{segdataset} dataset (8 videos chosen with a single moving object).
A parameter $\sigma$ (of Gaussian) controls the level
of jitter fused on the stable videos to create 72 synthetic jittery videos. The values for $\sigma$ are chosen
as 0.05, 0.15 and 0.25 for low, medium and high levels of jitter. The same settings of perturbation is uniformly used for all the methods. (App. A.3 of supplementary document~\cite{KendallSegSupp} has details of the method used to fuse realistic random jitter patterns, using homography estimated across frames).

The segmentation performance of the proposed method on SegTrack2~\cite{segdataset}, with different levels of jitter introduced, along with the state-of-the-art methods are shown in table~\ref{tab3}. The average IoU metric over all the videos in the dataset for each level of jitter are shown. On an average, our method outperform others for all the three categories of jitter (Visual results of both natural and synthetic jittery videos are given in the website~\cite{KendallSeg}).

To compare the performance of our method with the other state-of-the-art methods for stable videos, experiments were performed on four stable videos of Segtrack2~\cite{segdataset}, for which quoted numbers from the papers of the competing methods are taken. Table~\ref{stable} shows the comparison of four methods with the proposed method in stable videos. Average segmentation error (in pixels) measure is used to compare between the methods. Since the method~\cite{ochs2014segmentation} have not experimented on the videos from Segtrack2, the results are not quoted in the table. It can be seen that our method is comparable with the competing methods for all the four videos.

 \begin{table}[!h]
\begin{center}
 \begin{tabular}{ |p{2.3cm}|p{1.5cm}|p{1.5cm}|p{1.5cm}|p{1.5cm}| p{1.5cm}|}
\cline{2-6}
\multicolumn{1}{c|}{}&  \multicolumn{5}{c|}{\textbf{ Average segmentation error (in pixels) of competing methods}}\\ 
\hline
\textbf{Video}&~\cite{zhang2013video} & ~\cite{Papazoglou_2013_ICCV} & ~\cite{faktor2014video}&~\cite{wang2015saliency}&Proposed\\
\hline
Birdfall & 155&239&\textbf{146}&209&150 \\ 
\hline
Girl & 1488&2404&\textbf{797}&1040&1012 \\ \hline
Monkeydog & 365&306&361&562&\textbf{302} \\ \hline
Parachute & 220&347&219&207&\textbf{199} \\ \hline
\end{tabular}
\caption{Comparison of average segmentation error for four videos of SegTrack2 dataset~\cite{segdataset} quoting the reported numbers from the papers of the competing methods.}
\label{stable}
\end{center}
\end{table}

\textbf{Limitations}: Our method fails in the presence of independent uniform rigid motion (rotation or translation) throughout the frames. Such trajectories in Kendall's shape space are difficult to differentiate (refer to the website~\cite{KendallSeg} for observing the failure cases). Also, this method is experimented on a single moving object. Automatically identifying the number of motion segments and extending the proposed method to identify the independent motion segments with uniform rigid motion are the future directions of this work.

\section{Conclusion}
A novel trajectory clustering method for moving object segmentation in jittery videos based on analysis in Kendall's shape space, using Procrustes distance as a cue, has been proposed in the paper. Trajectories are modelled in Kendall's shape space and an iterative stabilization-clustering approach is followed for obtaining accurate foreground and background clusters. The stabilization of the trajectories is performed by Generalized Procrustes analysis (GPA) of trajectories. The stabilization improves the clustering accuracy and hence produces well formed clusters. Experiments were performed on 20 real-world natural jittery videos, for which the ground truth was manually annotated. Artificial jitter was fused on videos of a standard segmentation dataset. The performance analysis shows that the proposed method generates the best results when compared to all other state-of-the-art methods.

\textbf{Acknowledgment}- This work has been partially supported by TCS Foundation, India.

\end{document}